\documentclass{article}
\usepackage{spconf,amsmath,epsfig}

\title{COMPARISON OF CROSS-ENTROPY, DICE, AND FOCAL LOSS FOR SEA ICE TYPE SEGMENTATION}
\name{Rafael Pires de Lima$^1$, Behzad Vahedi$^1$, Morteza Karimzadeh$^1$\thanks{This research is supported by the National Science Foundation under Grant No. 2026962.}}
\address{$^1$ Department of Geography, University of Colorado Boulder}
\begin{document}
%
\maketitle
\begin{abstract}

\end{abstract}
Up-to-date sea ice charts are crucial for safer navigation in ice-infested waters. Recently, Convolutional Neural Network (CNN) models show the potential to accelerate the generation of ice maps for large regions. However, results from CNN models still need to undergo scrutiny as higher metrics performance not always translate to adequate outputs. Sea ice type classes are imbalanced, requiring special treatment during training. We evaluate how three different loss functions, some developed for imbalanced class problems, affect the performance of CNN models trained to predict the dominant ice type in Sentinel-1 images. Despite the fact that Dice and Focal loss produce higher metrics, results from cross-entropy seem generally more physically consistent. 

\section{Introduction}
\label{sec:intro}

The recent reduction in sea ice volume has led to increased potential for shipping in the Arctic (e.g., \cite{melia_sea_2016, pizzolato_influence_2016}), which requires improved marine information services to ensure both safe transit and minimal environmental impact. Several national ice centers periodically produce sea ice charts, used by navigators to find the best and safest shipping routes. The generation of ice charts continues heavily dependent on sea ice experts that manually interpret remotely sensed data in conjunction with other information to create maps that indicate ocean conditions. Machine learning was identified as a methodology with significant potential for accelerating the generation of sea ice charts. Convolutional Neural Networks (CNNs), in particular, showed promising results for sea ice characterization using remotely sensed data as input, in both image classification and semantic segmentation frameworks. In classification, subsampled patches of remotely sensed images are labeled as water or other sea ice attribute, and then patches are mosaiced into a classified scene (e.g., \cite{boulze_classification_2020, khaleghian_sea_2021}). Most studies approaching sea ice mapping as a semantic segmentation task have used U-Net \cite{ronneberger_u-net_2015} variations. Ren et al. \cite{ren_development_2021} added attention components into U-Net that further captured the spatial correlation of features in individual feature maps and the interrelation of pixel values between different feature maps. Adding attention components improved accuracy when compared to the original architecture for a binary segmentation of open water and sea ice. Wang and Li \cite{wang_arctic_2021} combined five U-Nets outputs to perform a binary segmentation of Sentinel-1 scenes into open water and sea ice. Stokholm et al. \cite{stokholm_ai4seaice_2022} explored expanding the number of convolutional blocks in a U-Net architecture to perform semantic segmentation of Sentinel-1 images to segment sea ice concentration. In \cite{boulze_classification_2020, khaleghian_sea_2021, ren_development_2021, wang_arctic_2021, stokholm_ai4seaice_2022}, the models were trained using categorical or binary cross-entropy (CE) loss. Wang and Li \cite{wang_arctic_2021} combined binary CE and Dice loss \cite{sudre_dice_2017} to train their models. Despite these improvements, automated ice type classification remains unreliable for operational purposes, which can be attributed to multiple potential reasons, including the subjectivity of sea ice expert interpretation (labels) or loss functions used to train models. Sea ice type is generally an imblanaced problem, and adopting different losses has the potential to improve the performance of the models without requiring significant changes in model development and training. Recently, Kucic and Stokholm \cite{Kucik2023} evaluated the effect of different losses for sea ice concentration mapping. Here, we use a semantic segmentation framework and rely on the ExtremeEarth V2 dataset \cite{hughes_extremeearth_2021} to perform a set of experiments and evaluate what is the effect in performance when we train the models using different loss functions for ice type mapping.

\begin{figure*}[h!]
\begin{minipage}[b]{1.0\linewidth}
 \centering
 \includegraphics[width=0.68\textwidth]{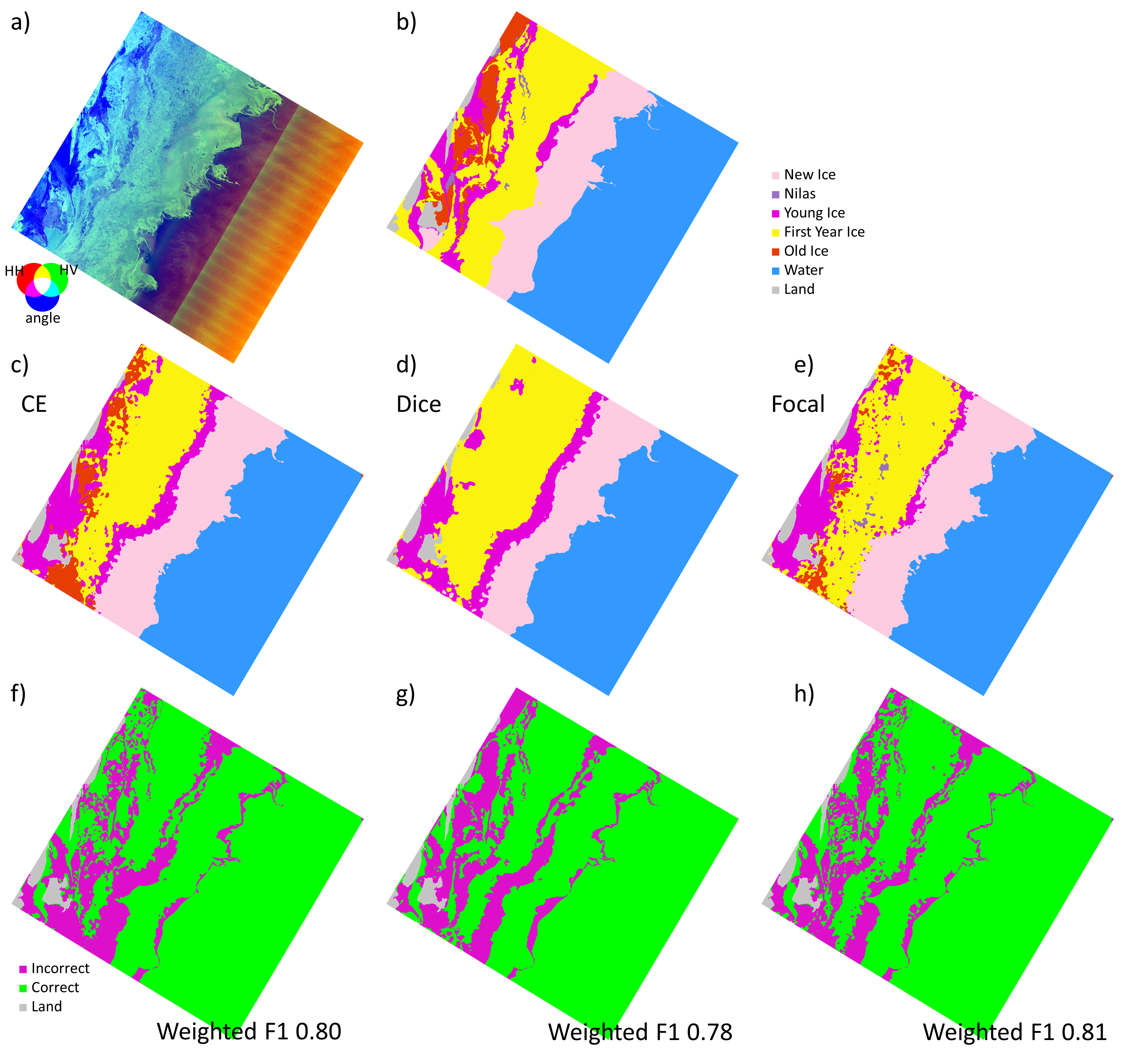}
\end{minipage}
\caption{Example of model predictions with highest scores for January. a) Input to the models, a Sentinel-1 image. b) Rasterized ExtremeEarth V2 labels. c), d), and e) share the same legend as b) and show predictions with highest weighted F1 for models trained with CE, Dice, and Focal loss, respectively. f), g), and h) share the same legend and show errors of losses compared to the labels in b). Although there is a larger mismatch between Dice results and ExtremeEarth V2 labels, the First Year Ice shows fewer blobs of other ice types in comparison to CE and Focal loss results.}
\label{fig1}
\end{figure*}

\section{Model architecture and training strategy}
\label{sec:model}

We approach sea ice type charting as a semantic segmentation task. To do so, we design a model architecture that uses the first three blocks of ResNet18 \cite{he_deep_2016} as encoder, and a decoder based on the Atrous Spatial Pyramid Pooling (ASPP) module \cite{chen_rethinking_2017}. The encoder and decoder sum 4 M trainable parameters. We initialize the encoder with ImageNet \cite{russakovsky_imagenet_2015} weights that are updated in training, while the decoder is randomly initialized. Our model takes as input Horizontal-Horizontal (HH), Horizontal-Vertical (HV), and incidence angle from Sentinel-1 images, and outputs ice type. We train the models using a batch size of 24, Adam optimizer \cite{kingma_adam_2014} with a learning rate starting at 1e-5. The learning rate is multiplied by 0.1 if the validation loss does not decrease in five epochs, to a minimum of 1e-8 and training stops if the validation loss does not decrease in 20 epochs. The weights with smallest validation loss are used for testing. These hyperparameters are kept the same, and we train models using three different loss functions with their default parameters: CE, Dice \cite{sudre_dice_2017}, and Focal \cite{lin_focal_2017}. We repeat each experiment three times to account for stochastic variation.  

\section{Dataset pre-processing}
\label{sec:data}

The ExtremeEarth V2 dataset \cite{hughes_extremeearth_2021} is a collection of labels in the form of high-resolution ice charts for twelve Sentinel-1 images in Extra Wide (EW) mode acquired over the East Coast of Greenland. The twelve images are roughly one month apart, one for each month of 2018, and show several types of sea ice under different weather conditions. ExtremeEarth V2 labels are polygons representing sea ice or open water with different characteristics. Each polygon contains attributes characterizing oldest sea ice, second oldest sea ice, as well as their respective concentrations. We derive the “dominant ice type” by computing the sea ice with the highest concentration in a polygon and use that as the target for our CNN models. We separate two scenes, January and July for testing. We clip half of February, June, August, and December for validation. Test and validation outputs are generated on a single pass for full images. To generate training samples, we extract 100 randomly placed patches of size 80 km$^2$, which correspond to 1000 x 1000 pixels of Sentinel-1 images that were resampled to 80 x 80 m pixel size. 

\section{Results and Discussion}
\label{sec:res-and-disc}
\begin{figure*}[htb]
\begin{minipage}[b]{1.0\linewidth}
 \centering
 \includegraphics[width=0.68\textwidth]{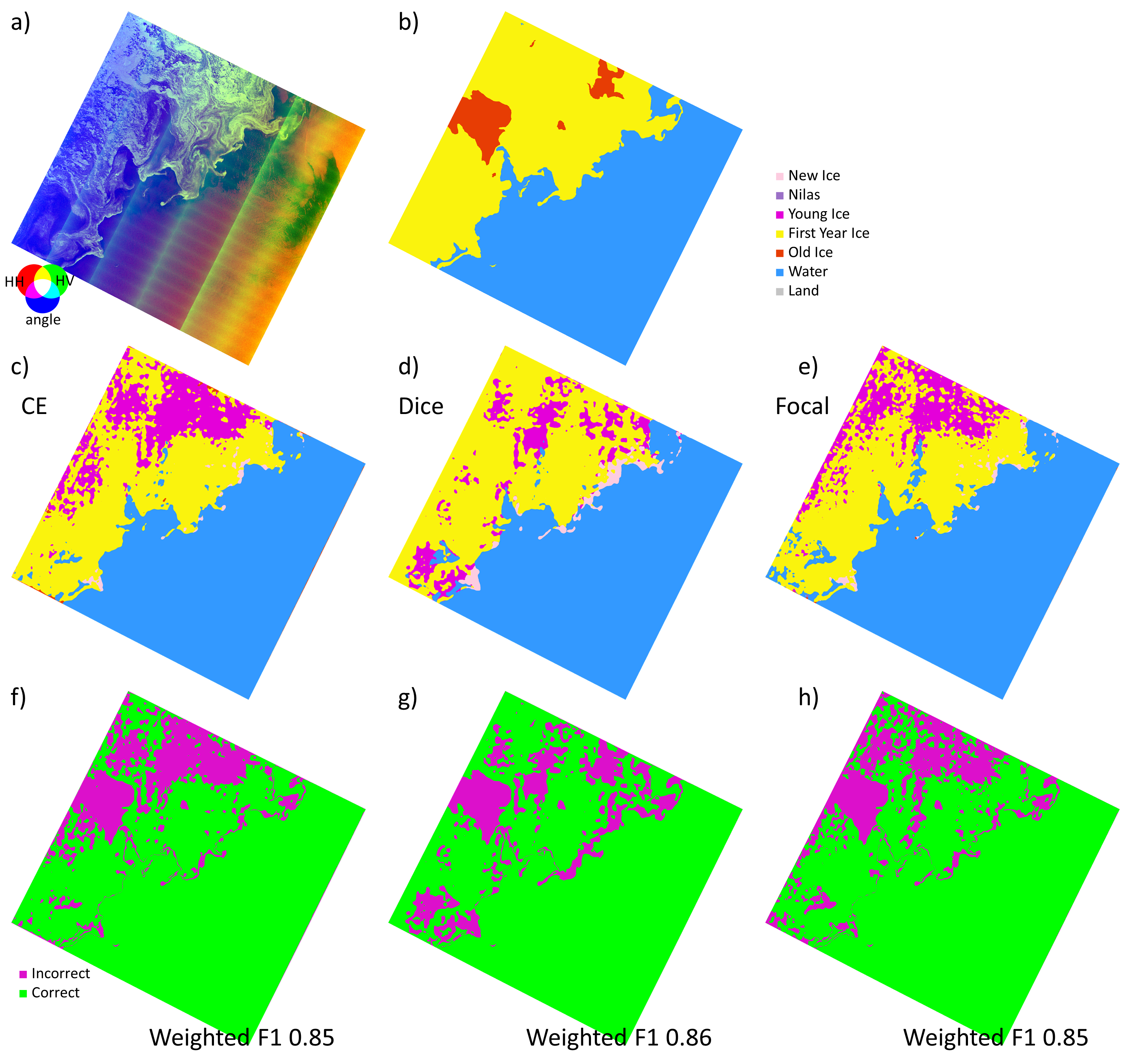}
\end{minipage}
\caption{Example of model predictions with highest scores for July. a) Input to the models, a Sentinel-1 image. b) Rasterized ExtremeEarth V2 labels. c), d), and e) share the same legend as b) and show predictions with highest weighted F1 for models trained with CE, Dice, and Focal loss, respectively. f), g), and h) share the same legend and show errors of losses compared to the labels in b). }
\label{fig2}
\end{figure*}

Results show similar performance metrics for all losses investigated, with a marginal improvement for Focal loss. The average test weighted F1 score (average across January and July for all three experiments) for CE is 0.815 (minimum of 0.805, maximum of 0.825). Dice results are 0.810 (0.80, 0.82), and Focal results are 0.822 (0.81, 0.83). Although Dice and Focal loss were developed for imbalanced datasets, our results show comparable performance for the more common CE loss for the ExtremeEarth V2 dataset. This potentially hints at other reasons that might have contributed to a relatively low performance in ice type classification, including lack of sufficient training samples, subjective labels at the polygon level as interpreted by the sea ice analyst, and ambiguity of SAR backscatter over different types.
However, evaluating the results solely based on metrics performance can be problematic. Fig. \ref{fig1} shows the results for the models with the highest weighted F1 for January. The model trained with Dice loss misses all old ice and classifies those regions as first-year ice. Due to the difference in thickness, incorrectly identifying old ice as first-year ice can have serious consequences for shipping and other marine operations. Of the three proposed methods, CE results appear to best capture the old ice areas. All models struggle to correctly identify the ice edge, which is attributed to a combination of the model's architecture resolution and the fact that the new ice at the edge appears darker in the SAR image.
Fig. \ref{fig2} shows results for the models with the highest weighted F1 for July, during Summer when ice is generally melting and there is no ice growth. All models miss the old ice regions that are classified either as young or first-year ice. The weighted F1 for the model trained with Dice is the highest, however, the model predicts relatively large new ice regions when compared to CE or Focal, an undesirable outcome. 

\section{Conclusions}
\label{sec:conclusions}

We evaluated how different loss functions affect the semantic segmentation of CNN models trained to classify the dominant ice type of ExtremeEarth V2 labels. Although Dice and Focal loss sometimes show better performance metrics, visual inspection indicates that predictions from models trained with CE loss are generally more physically reasonable. Further research for proper classification of minority classes might help in the loss function choice. 

\section{Acknowledgments}
\label{sec:acknow}
This material is based upon work supported by the National Science Foundation under Grant No. 2026962. We thank the Extreme Earth project and MET Norway for making the ExtremeEarth dataset available to the sea ice community. The code used for this research is available at https://github.com/geohai/sea-ice-segment.





\bibliography{bibtex/bib/IEEEabrv.bib,bibtex/bib/main.bib}{}

\begin{thebibliography}{10}
\providecommand{\url}[1]{#1}
\csname url@samestyle\endcsname
\providecommand{\newblock}{\relax}
\providecommand{\bibinfo}[2]{#2}
\providecommand{\BIBentrySTDinterwordspacing}{\spaceskip=0pt\relax}
\providecommand{\BIBentryALTinterwordstretchfactor}{4}
\providecommand{\BIBentryALTinterwordspacing}{\spaceskip=\fontdimen2\font plus
\BIBentryALTinterwordstretchfactor\fontdimen3\font minus
  \fontdimen4\font\relax}
\providecommand{\BIBforeignlanguage}[2]{{%
\expandafter\ifx\csname l@#1\endcsname\relax
\typeout{** WARNING: IEEEtran.bst: No hyphenation pattern has been}%
\typeout{** loaded for the language `#1'. Using the pattern for}%
\typeout{** the default language instead.}%
\else
\language=\csname l@#1\endcsname
\fi
#2}}
\providecommand{\BIBdecl}{\relax}
\BIBdecl

\bibitem{melia_sea_2016}
N.~Melia, K.~Haines, and E.~Hawkins, ``\BIBforeignlanguage{en}{Sea ice decline
  and 21st century trans-{Arctic} shipping routes},''
  \emph{\BIBforeignlanguage{en}{Geophysical Research Letters}}, vol.~43,
  no.~18, pp. 9720--9728, 2016.

\bibitem{pizzolato_influence_2016}
L.~Pizzolato, S.~E.~L. Howell, J.~Dawson, F.~Laliberté, and L.~Copland,
  ``\BIBforeignlanguage{en}{The influence of declining sea ice on shipping
  activity in the {Canadian} {Arctic}},''
  \emph{\BIBforeignlanguage{en}{Geophysical Research Letters}}, vol.~43,
  no.~23, pp. 12,146--12,154, 2016.

\bibitem{boulze_classification_2020}
H.~Boulze, A.~Korosov, and J.~Brajard, ``\BIBforeignlanguage{en}{Classification
  of {Sea} {Ice} {Types} in {Sentinel}-1 {SAR} {Data} {Using} {Convolutional}
  {Neural} {Networks}},'' \emph{\BIBforeignlanguage{en}{Remote Sensing}},
  vol.~12, no.~13, p. 2165, Jan. 2020, number: 13 Publisher: Multidisciplinary
  Digital Publishing Institute.

\bibitem{khaleghian_sea_2021}
S.~Khaleghian, H.~Ullah, T.~Kræmer, N.~Hughes, T.~Eltoft, and A.~Marinoni,
  ``\BIBforeignlanguage{en}{Sea {Ice} {Classification} of {SAR} {Imagery}
  {Based} on {Convolution} {Neural} {Networks}},''
  \emph{\BIBforeignlanguage{en}{Remote Sensing}}, vol.~13, no.~9, p. 1734, Jan.
  2021, number: 9 Publisher: Multidisciplinary Digital Publishing Institute.

\bibitem{ronneberger_u-net_2015}
O.~Ronneberger, P.~Fischer, and T.~Brox, ``U-{Net}: {Convolutional} {Networks}
  for {Biomedical} {Image} {Segmentation},'' in \emph{Medical {Image}
  {Computing} and {Computer}-{Assisted} {Intervention} – {MICCAI} 2015},
  N.~Navab, J.~Hornegger, W.~M. Wells, and A.~F. Frangi, Eds.\hskip 1em plus
  0.5em minus 0.4em\relax Cham: Springer International Publishing, 2015, pp.
  234--241.

\bibitem{ren_development_2021}
Y.~Ren, X.~Li, X.~Yang, and H.~Xu, ``Development of a {Dual}-{Attention}
  {U}-{Net} {Model} for {Sea} {Ice} and {Open} {Water} {Classification} on
  {SAR} {Images},'' \emph{IEEE Geoscience and Remote Sensing Letters}, vol.~19,
  pp. 1--5, 2021, conference Name: IEEE Geoscience and Remote Sensing Letters.

\bibitem{wang_arctic_2021}
Y.-R. Wang and X.-M. Li, ``\BIBforeignlanguage{English}{Arctic sea ice cover
  data from spaceborne synthetic aperture radar by deep learning},''
  \emph{\BIBforeignlanguage{English}{Earth System Science Data}}, vol.~13,
  no.~6, pp. 2723--2742, Jun. 2021, publisher: Copernicus GmbH.

\bibitem{stokholm_ai4seaice_2022}
A.~Stokholm, T.~Wulf, A.~Kucik, R.~Saldo, J.~Buus-Hinkler, and S.~M.
  Hvidegaard, ``{AI4SeaIce}: {Toward} {Solving} {Ambiguous} {SAR} {Textures} in
  {Convolutional} {Neural} {Networks} for {Automatic} {Sea} {Ice}
  {Concentration} {Charting},'' \emph{IEEE Transactions on Geoscience and
  Remote Sensing}, vol.~60, pp. 1--13, 2022.

\bibitem{sudre_dice_2017}
C.~H. Sudre, W.~Li, T.~Vercauteren, S.~Ourselin, and M.~Jorge~Cardoso,
  ``\BIBforeignlanguage{en}{Generalised {Dice} {Overlap} as a {Deep} {Learning}
  {Loss} {Function} for {Highly} {Unbalanced} {Segmentations}},'' in
  \emph{\BIBforeignlanguage{en}{Deep {Learning} in {Medical} {Image} {Analysis}
  and {Multimodal} {Learning} for {Clinical} {Decision} {Support}}}, ser.
  Lecture {Notes} in {Computer} {Science}, M.~J. Cardoso, T.~Arbel,
  G.~Carneiro, T.~Syeda-Mahmood, J.~M.~R. Tavares, M.~Moradi, A.~Bradley,
  H.~Greenspan, J.~P. Papa, A.~Madabhushi, J.~C. Nascimento, J.~S. Cardoso,
  V.~Belagiannis, and Z.~Lu, Eds.\hskip 1em plus 0.5em minus 0.4em\relax Cham:
  Springer International Publishing, 2017, pp. 240--248.

\bibitem{Kucik2023}
\BIBentryALTinterwordspacing
A.~Kucik and A.~Stokholm, ``{AI}4seaice: selecting loss functions for automated
  {SAR} sea ice concentration charting,'' \emph{Scientific Reports}, vol.~13,
  no.~1, Apr. 2023. [Online]. Available:
  \url{https://doi.org/10.1038/s41598-023-32467-x}
\BIBentrySTDinterwordspacing

\bibitem{hughes_extremeearth_2021}
\BIBentryALTinterwordspacing
N.~Hughes and F.~Amdal, ``{ExtremeEarth} {Polar} {Use} {Case} {Training}
  {Data},'' 2021, publisher: Zenodo Version Number: 1.0.0. [Online]. Available:
  \url{https://doi.org/10.5281/zenodo.4683174}
\BIBentrySTDinterwordspacing

\bibitem{he_deep_2016}
\BIBentryALTinterwordspacing
K.~He, X.~Zhang, S.~Ren, and J.~Sun, ``Deep {Residual} {Learning} for {Image}
  {Recognition},'' in \emph{2016 {IEEE} {Conference} on {Computer} {Vision} and
  {Pattern} {Recognition} ({CVPR})}.\hskip 1em plus 0.5em minus 0.4em\relax
  IEEE, Jun. 2016, pp. 770--778. [Online]. Available:
  \url{http://ieeexplore.ieee.org/document/7780459/}
\BIBentrySTDinterwordspacing

\bibitem{chen_rethinking_2017}
L.-C. Chen, G.~Papandreou, F.~Schroff, and H.~Adam, ``Rethinking {Atrous}
  {Convolution} for {Semantic} {Image} {Segmentation},'' \emph{arXiv e-prints},
  p. arXiv:1706.05587, Jun. 2017, \_eprint: 1706.05587.

\bibitem{russakovsky_imagenet_2015}
O.~Russakovsky, J.~Deng, H.~Su, J.~Krause, S.~Satheesh, S.~Ma, Z.~Huang,
  A.~Karpathy, A.~Khosla, M.~Bernstein, A.~C. Berg, and L.~Fei-Fei,
  ``{ImageNet} {Large} {Scale} {Visual} {Recognition} {Challenge},''
  \emph{International Journal of Computer Vision}, vol. 115, no.~3, pp.
  211--252, Dec. 2015, publisher: Springer US.

\bibitem{kingma_adam_2014}
D.~P. Kingma and J.~Ba, ``Adam: {A} {Method} for {Stochastic} {Optimization},''
  in \emph{{ICLR}}.\hskip 1em plus 0.5em minus 0.4em\relax San Diego, CA, USA:
  ICLR, Dec. 2014, p. arXiv:1412.6980.

\bibitem{lin_focal_2017}
T.-Y. Lin, P.~Goyal, R.~Girshick, K.~He, and P.~Dollár, ``Focal {Loss} for
  {Dense} {Object} {Detection},'' in \emph{2017 {IEEE} {International}
  {Conference} on {Computer} {Vision} ({ICCV})}, Venice, Italy, Oct. 2017, pp.
  2999--3007, iSSN: 2380-7504.

\end{thebibliography}
\bibliographystyle{IEEEtran}
\end{document}